%% file: main.tex
\documentclass[10pt, a4paper]{article}

\usepackage[final]{lrec2026} % this is the new style
\usepackage[table]{xcolor}
\usepackage{listings}

\definecolor{lightgrayrow}{gray}{0.97} 
\definecolor{highlightBlue}{RGB}{93, 173, 226}

\definecolor{highlightYellow}{RGB}{118, 146, 255}
\definecolor{highlightGreen}{RGB}{117, 72, 94}
\definecolor{highlightRed}{RGB}{203, 144, 77}
\usepackage{amssymb}
\usepackage{amsmath}
\usepackage{booktabs}   % Für schöne Tabellenlinien
\usepackage{array}  
\usepackage{microtype}  % Für bessere typografische Abstände
\usepackage{xcolor}
\usepackage{multirow}
\usepackage{makecell}
\usepackage{subcaption}
\usepackage{url}
\usepackage{rotating}
\usepackage{enumitem} 
\usepackage{tikz}
\usepackage{pgfplots}
\pgfplotsset{compat=1.18}
\usepackage{placeins}
\usepackage{float}
\usepackage{xurl}

\title{Annotation Quality in Aspect-Based Sentiment Analysis: 
A Case Study Comparing Experts, Students, Crowdworkers, and Large Language Models}

\name{
Niklas Donhauser$^{1}$ \quad
Jakob Fehle$^{1}$ \quad
Nils Constantin Hellwig$^{1}$ 
\\
\textbf{Markus Weinberger}$^{1}$ \quad
\textbf{Udo Kruschwitz}$^{2}$ \quad
\textbf{Christian Wolff}$^{1}$ \\[0.5em]
}
\address{$^{1}$ Media Informatics Group, University of Regensburg, Regensburg, Germany \\
$^{2}$ Information Science Group, University of Regensburg, Regensburg, Germany\\
\texttt{\{niklas.donhauser,jakob.fehle,nils-constantin.hellwig,} \\
\texttt{markus.weinberger,udo.kruschwitz,christian.wolff\}@ur.de}}

\input{chapters/abstract}

\begin{document}

\maketitleabstract
%-----------------------------Chapters-----------------------------
\section{Introduction}
\input{chapters/introduction}
\section{Related Work}

\input{chapters/related_work}

\section{Methodology}
\input{chapters/methodology}
\section{Experiments}
\input{chapters/experiments}
\section{Results and Discussion}
\input{chapters/results_discussion}
\section{Conclusion}
\input{chapters/conclusion}

\section*{Limitations}
\input{chapters/limitations}
\section*{Ethical Considerations}
\input{chapters/ethical_considerations}
%-----------------------------ONLY CAMERA READY-----------------------------
%\section*{Acknowledgments}
%\input{chapters/acknowledgments}

%citation for work and Language Ressources
%\citet{Eco:1990}
%\citetlanguageresource{Speecon}

%-----------------------------Reference-----------------------------
\section{Bibliographical References}\label{sec:reference}

\bibliographystyle{lrec2026-natbib}
\bibliography{main}
% \bibliography{x_ref}

% \section{Language Resource References}
% \bibliographystylelanguageresource{lrec2026-natbib}
% \bibliographylanguageresource{x_languageresource}

%-----------------------------Appendix ONLY CAMERA READY-----------------------------
\appendix
\onecolumn
\section*{Appendix}
\section{Prompts for Few-Shot LLMs}
\label{App:prompts_for_fs_LLMS}
\input{appendix/prompts}
\section{Annotation Examples}
\label{App:annotation_example}
\input{appendix/annotation_example}
\newpage
\input{appendix/4_annotations_example}
% \section{Training Time and Memory Usage}
% \input{appendix/training_time_and_memory_usage}
\section{Dataset Statistics}
\label{App:dataset_statistics}
\input{appendix/sentiment_dataset_statistics}

\input{appendix/category_dataset_statistics}
\newpage
\section{Questionnaire Results}
\label{App:questionaire_results}
\input{appendix/questionaire_results}
\newpage
\section{Label Interface}
\label{App:label_interface}
\input{appendix/label_interface}
\end{document}

%% file: chapters/abstract.tex
\abstract{Aspect-Based Sentiment Analysis (ABSA) enables fine-grained opinion analysis by identifying sentiments toward specific aspects or targets within a text.
While ABSA has been widely studied for English, research on other languages such as German remains limited, largely due to the lack of high-quality annotated datasets.
This paper examines how different annotation sources influence the development of German ABSA. 
To this end, an existing dataset is re-annotated by experts to establish a ground truth, which serves as a reference for evaluating annotations produced by students, crowdworkers, Large Language Models (LLMs), and experts. 
Annotation quality is compared using Inter-Annotator Agreement (IAA) and its impact on downstream model performance for different ABSA subtasks.
The evaluation focuses on Aspect Category Sentiment Analysis (ACSA) and Target Aspect Sentiment Detection (TASD). 
We apply State-of-the-Art (SOTA) methods for ABSA, including BERT-, T5-, and LLaMA-based approaches to assess performance differences, spanning fine-tuning and in-context learning with instruction prompts. 
The findings provide practical insights into trade-offs between annotation reliability, and efficiency, offering guidance for dataset construction in under-resourced Natural Language Processing (NLP) scenarios.
\\ 
\newline 
\Keywords{Aspect-Based Sentiment Analysis, Large Language Models, Restaurant Reviews, Annotation Quality, Dataset Annotation} 
}

%% file: chapters/introduction.tex
%The growing availability of user-generated content such as product reviews, customer feedback, and social media posts has made sentiment analysis one of the most widely studied tasks in natural language processing (NLP) \citep{wankhade_survey_2024, brauwers_survey_2022, chauhan_aspect_2023}. 
%Sentiment analysis aims to automatically identify and classify subjective opinions in texts, thereby enabling the extraction of valuable insights from large volumes of unstructured data.

%While traditional sentiment analysis focuses on the overall polarity of a text, aspect-based sentiment analysis (ABSA) extends this perspective by examining sentiments tied to specific aspects or attributes \citep{liu_sentiment_2022}. 
%This finer granularity offers deeper insights, making ABSA particularly useful in domains where user opinions cover multiple dimensions.
Aspect-Based Sentiment Analysis (ABSA) is a subfield of Natural Language Processing (NLP) concerned with identifying sentiment expressed toward specific aspects or attributes mentioned in text.
With the rapid growth of online content, user-generated data has become a major source for understanding public opinion across a wide range of domains \citep{liu_sentiment_2022}. 
ABSA has been applied to product and restaurant reviews, movie critiques, political discourse, educational initiatives, social events, and market campaigns, as well as government policy analysis \citep{hua_systematic_2024}.
It has also been used in social media contexts such as YouTube video ranking and in the economic domain, where aspect-level models are applied to microblogs and news articles \citep{chauhan_aspect_2023}.

Established benchmark corpora have played a central role in shaping ABSA research. 
The SemEval shared tasks from 2014 to 2016 defined standard datasets and evaluation settings that have strongly influenced subsequent work \citep{pontiki_semeval-2014_2014, pontiki_semeval-2015_2015, pontiki_semeval-2016_2016, chebolu_review_2023}. 
Within these benchmarks, the restaurant domain emerged as one of the most prominent and widely reused settings, and has since become a standard benchmark setting for ABSA across multiple languages.
Beyond its methodological importance, the restaurant domain also carries clear practical relevance, as aspect-level sentiment information can support applications such as recommendation systems and customer feedback analysis \citep{ara_understanding_2020, singhi_exploring_2024}.
%Restaurant reviews are especially suitable for this purpose: they are widely used as benchmark domains in ABSA research across many languages \citep{pontiki_semeval-2014_2014, pontiki_semeval-2015_2015, pontiki_semeval-2016_2016, chebolu_review_2023} and they also have clear practical relevance for applications like recommendation systems and customer feedback analysis \citep{ara_understanding_2020, singhi_exploring_2024}.

Despite the potential of ABSA, its success strongly depends on the availability of annotated training data. 
However, ABSA is an under-resourced task for many languages, including German: datasets are scarce, and existing resources are often limited in size, domain coverage, or annotation quality \citep{fehle_aspect-based_2023, hellwig_gerestaurant_2024}.
Constructing high-quality datasets requires careful annotation, but this process is costly, time-intensive, and prone to inconsistencies \citep{klie_analyzing_2024, monarch_human---loop_2021, orr_social_2024, dobnik_role_2023}.
The challenge is amplified by the fact that different annotation strategies, such as crowdsourcing \citep{nowak2010reliable, he_if_2024_crowdworker}, student annotators \citep{fehle_aspect-based_2023}, expert annotators \citep{fehle_german_2025, barbarestani_content_2024}, or the use of Large Language Models (LLMs) \citep{ostyakova-etal-2023-chatgpt, Hellwig2025-sd, maehlum_its_2024} can produce datasets of varying reliability and utility for machine learning models.
%\citep{ostyakova-etal-2023-chatgpt} crwodsourcing (chatgpt and experts)
%\citep{nowak2010reliable} better crowsourcing paper probably
%\citep{hellwig_we_2025} LMM annotation
%\citep{fehle_aspect-based_2023} student annotators
%\citep{klie_analyzing_2024}

This raises the question of how annotation strategies influence downstream ABSA performance and whether higher annotation quality justifies increased effort. 
To address this, we conduct a systematic comparison of four annotator groups, (1) crowdworkers, (2) students, (3) LLMs, and (4) task experts, in the German restaurant review domain. 
We evaluate the resulting datasets on two central ABSA subtasks, Aspect Category Sentiment Analysis (ACSA) and Target Aspect Sentiment Detection (TASD) \citep{fehle_german_2025, Hellwig2025-sd, bu_asap_2021, wu_m-absa_2025}, and complement model-based evaluation with Inter-Annotator Agreement (IAA) to assess annotation consistency.

% This raises the central question of how datasets based on those annotations influence the performance of ABSA systems and whether the benefits of high-quality annotations (e.g. by experts) justify their associated costs.

% We therefore conduct a systematic comparison of annotation strategies and assess their influence on ABSA performance with first experiments in the German restaurant review domain.

% Specifically, the study investigates annotations produced by four groups: (1) crowdworkers, (2) students, (3) LLMs, and (4) task experts.
% \begin{itemize}[itemsep=0pt, topsep=2pt]
%     \item Crowdworkers
%     \item Students
%     \item LLMs
%     \item Task Experts
% \end{itemize}
% By comparing these approaches, the paper seeks to uncover quality differences in the resulting datasets and to examine their consequences for machine learning performance. 
% The evaluation concentrates on two key subtasks of ABSA: Aspect Category Sentiment Analysis (ACSA) and Target Aspect Sentiment Detection (TASD) \citep{fehle_german_2025, Hellwig2025-sd, bu_asap_2021, wu_m-absa_2025}. 

In addition to evaluating model performance, the study also analyzes IAA as a complementary measure to assess annotation consistency across different annotator groups.
To comprehensively assess the influence of annotation quality, a range of State-of-the-Art (SOTA) approaches are applied and compared. 
These include traditional classifier-based models such as BERT-CLF~\cite{fehle_aspect-based_2023}, HIER-GCN~\cite{Cai2020-km}, as well as more recent text generation and LLM techniques, including Paraphrase~\cite{zhang_method_paraphrase}, Multi-View Prompting~\cite{gou_method_mvp}, or LLaMA~\cite{Dubey2024-pu} with fine-tuning and Gemma~\cite{gemmateam2025gemma3technicalreport} with few-shot prompting. 

% In summary, this paper conducts a systematic annotation study on German restaurant reviews, comparing four different annotation strategies: crowdworkers, students, large language models, and experts. 
% This work evaluates their impact on the quality of ABSA datasets and examines how annotation differences influence model performance across two key subtasks: ACSA and TASD.
% By doing so, it provides both practical recommendations for dataset construction and theoretical insights into the role of annotation quality in NLP.
% To support reproducibility and further research, all code is provided on GitHub,\footnote{GitHub: Removed for review} while the datasets are available upon request to ensure responsible usage for academic purposes and to preserve the original intent of the dataset.

In summary, this paper presents a systematic annotation study on German restaurant reviews, comparing four annotation strategies: crowdworkers, students, LLMs, and experts, and analyzes their impact on ABSA dataset quality and downstream model performance for ACSA and TASD. The study derives practical recommendations for dataset construction and provides empirical insights into how annotation quality affects ABSA models. To support reproducibility, the code is publicly available on GitHub,\footnote{\scriptsize GitHub:\,\hangindent=2.3em\hangafter=1 \href{https://github.com/NiklasDonhauser/absa-annotation-quality}
{https://github.com/NiklasDonhauser/\\absa-annotation-quality}} and the datasets can be accessed upon request for academic use.

%% file: chapters/related_work.tex
A persistent challenge in ABSA research concerns the limited availability, diversity, and transparency of annotated datasets \citep{hua_systematic_2024}. Existing benchmarks are heavily concentrated in a small number of English review domains, most prominently the SemEval Restaurant and Laptop datasets. While these resources have driven methodological progress, they represent comparatively simplified settings and often yield inflated performance estimates on narrow domain slices \citep{hua_systematic_2024}. 

Beyond dataset size, annotation quality and documentation constitute a second critical bottleneck. Modern ABSA formulations such as triplet or quadruplet annotations translate directly into concrete dataset requirements, including a clearly specified label space and guidelines, trained annotators, annotation tools that support the intended output format, and transparent procedures for assessing annotation quality \citep{pontiki_semeval-2016_2016,klie_inception_2018}.  Meeting these requirements is time- and cost-intensive. However, many ABSA datasets provide only limited information about sampling strategies, annotator backgrounds, agreement measures, or conflict resolution procedures, complicating reproducibility and reliability assessment. Meta-analyses of NLP datasets highlight recurring deficiencies along dimensions such as stability, reproducibility, accuracy, and unbiasedness \citep{klie_analyzing_2024}. 

These structural issues become even more pronounced in non-English settings, where data scarcity and domain concentration further restrict systematic comparison and reuse.

\subsection{The State of ABSA Annotation in German}
Against this background, German provides a representative case to examine how structural challenges of ABSA dataset construction materialize in a non-English context. In contrast to English, where shared tasks and benchmark consolidation have shaped methodological development, German ABSA resources have emerged in a more fragmented manner, varying substantially in domain coverage, annotation granularity, and accessibility.
% Mapping these datasets allows us to assess not only their size and domain focus, but also how annotation design decisions affect comparability, reuse, and robustness across studies.
The majority of German ABSA datasets provide sentence-level annotations, including \textit{Hotel Reviews} \citep{fehle_aspect-based_2023}, \textit{GERestaurant} \citep{hellwig_gerestaurant_2024}, \textit{MobASA} \citep{gabryszak_mobasa_2022}, \textit{Talk of Literature} \citep{greve_aspect-based_2021}, and \textit{B2B Software Reviews} \citep{ fehle_german_2025}. 
Review-level annotations are offered by \textit{GermEval 2017} \citep{Wojatzki2017-gp}, while \textit{M-ABSA} provides sentence-level German data via automatic translation, lacking human-authored annotations and ground truth \citep{wu_m-absa_2025}. 
Access to several datasets is restricted, as some are proprietary or require direct author contact.

\subsection{Annotation Practices in the Literature}
%-----------------------------Guidelines-----------------------------

Clear annotation guidelines are essential for consistency in NLP tasks \citep{klie_analyzing_2024}. They define objectives, label spaces, and decision rules, and are often refined iteratively. Their structure can influence annotator behavior and introduce biases, making careful design crucial.
In ABSA, the SemEval shared tasks (2014–2016) established standardized task definitions and guidelines \citep{pontiki_semeval-2014_2014, pontiki_semeval-2015_2015, pontiki_semeval-2016_2016}, which have been widely reused and adapted. 
For German, GermEval 2017 followed a similar structure with stronger emphasis on practical instructions and language-specific phenomena \citep{Wojatzki2017-gp}, and subsequent German datasets build on these principles \citep{hellwig_gerestaurant_2024, fehle_german_2025}.

IAA is often reported as an indicator of annotation consistency \citep{klie_analyzing_2024}, but it should not be interpreted as a complete measure of annotation quality.
%IAA is commonly used to assess annotation quality \citep{klie_analyzing_2024}. 
In ABSA, span-based and structured annotations complicate agreement measurement. 
While Krippendorff’s $\alpha$ is often applied, many studies report F1-based agreement scores due to the difficulty of using chance-corrected metrics for span extraction \citep{pontiki_semeval-2016_2016, chebolu_review_2023}. 
However, agreement alone is insufficient; prior work recommends complementary quality-control measures, such as manual inspection or control instances, and distinguishes between intrinsic evaluation (e.g., IAA) and extrinsic evaluation via downstream performance for application use cases \citep{klie_analyzing_2024, jurafsky3}.

%% file: chapters/methodology.tex
This section describes the annotation and evaluation methodology. 
First, an independent ground truth was constructed to serve as a reference for evaluation. 
Subsequently, four distinct annotation settings were implemented, involving crowdworkers, students, LLMs, and task experts. Annotation was performed in batches of 200 sentences, while different mechanisms were used to ensure annotation quality (majority vote, curation, iterative refinement). 
This design enables a systematic comparison of annotation styles, their interrater agreement, and their impact on downstream model performance.

We build upon GERestaurant \cite{hellwig_gerestaurant_2024}, a German ABSA dataset available upon request. 
It contains 2,154 training and 924 test instances. 
We use the full test split as ground truth and randomly sample 1,000 training sentences due to annotation resource constraints.
The restaurant domain is chosen as it constitutes a widely established benchmark setting for ABSA across languages.

% Annotation Task %

\subsection{Annotation Objective}

The objective of our annotation studies is to identify all aspect–sentiment pairs expressed within a sentence, following the standard definition of ABSA. Each sentence constitutes one annotation unit, and multiple aspects per sentence are explicitly allowed.

This work focuses on the two established ABSA tasks in the literature: Aspect Category Sentiment Analysis (ACSA) and Target Aspect Sentiment Detection (TASD) \citep{zhang_survey_2023, chebolu_review_2023}. In both tasks, annotators assign one or more predefined aspect categories — \texttt{FOOD}, \texttt{SERVICE}, \texttt{AMBIENCE}, \texttt{PRICE}, and \texttt{GENERAL} — together with a sentiment polarity label — \texttt{POSITIVE}, \texttt{NEGATIVE}, \texttt{NEUTRAL}, and \texttt{CONFLICT}. The category \texttt{GENERAL} captures overall evaluations of the restaurant that cannot be attributed to one of the other four categories. \texttt{CONFLICT} is used when opposing sentiments toward the same aspect (or aspect phrase for TASD) are expressed within a sentence.

\begin{table}[ht]
\centering
\includegraphics[width=0.5\textwidth]{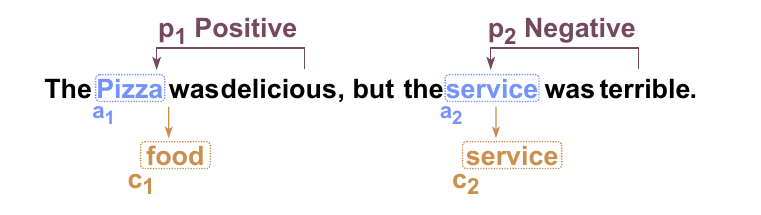}
\begin{tabular}{p{5,0cm}|c}
\toprule
\textbf{Subtask} & \textbf{Output} \\
\midrule
{\scriptsize Aspect Category Sentiment Analysis (ACSA)} & (\textcolor{highlightRed}{c}, \textcolor{highlightGreen}{p})\\
{\scriptsize Target Aspect Sentiment Detection (TASD)} & (\textcolor{highlightYellow}{a}, \textcolor{highlightRed}{c}, \textcolor{highlightGreen}{p})\\
\bottomrule
\end{tabular}
\caption{Illustration of the ABSA subtasks investigated in this study. Figure based on \citet{fehle_german_2025}.}
\label{fig:absa_example}
\end{table}

Implicit aspects must be identified in both tasks, i.e., cases where a sentiment clearly refers to a predefined category without explicitly mentioning it (e.g., “It tasted really good.” $\rightarrow$ \texttt{FOOD}). 
As depicted in Table~\ref{fig:absa_example}, the tasks differ in their annotation granularity: ACSA requires aspect category and sentiment polarity assignment only, whereas TASD additionally requires annotating the textual span that realizes the aspect category. 
For implicit aspects in TASD, no text span is marked.
An annotated example from the test set is provided in Appendix~\ref{App:annotation_example}.

%
%-----------------------------Annotation Strategies-----------------------------
%
\subsection{Annotation Strategies}
In addition to a new \textbf{expert-based ground truth}, we applied four different annotation strategies to construct ABSA datasets: \textbf{crowdworkers}, \textbf{students}, \textbf{LLMs}, and \textbf{experts}. 
%The first four strategies are described below, while the ground truth annotation is treated separately, as it serves as the evaluation benchmark for assessing consistency, reliability, and accuracy across strategies.

Due to limited expert availability, for both ground truth and expert anntoation, no separate dataset variant was created for ACSA.
Instead, the ACSA dataset was derived from the revised TASD annotations by removing aspect phrases and consolidating duplicate tuples.

%-----------------------------GroundTruth-----------------------------
\paragraph{Ground Truth} The ground truth dataset was annotated following the guidelines of \citet{klie_analyzing_2024}. 
It comprises 924 sentences annotated independently by two annotators with prior ABSA experience. 

%After each batch, inter-annotator agreement was computed using micro-averaged F1 \citep{chebolu_review_2023}, and disagreements were jointly reviewed.
To enable incremental quality control, the data were divided into five batches of approximately 185 sentences, after each of which IAA was computed using micro-averaged F1. 
Following \citet{chebolu_review_2023}, we prefer F1 over chance-corrected measures such as Kappa $\kappa$ for span extraction tasks. 
Disagreements were jointly reviewed, and recurring disagreement patterns informed iterative refinements of the annotation guidelines to reduce ambiguity and improve consistency.
These refinements mainly concerned the scope and specificity of valid aspect phrases, including the treatment of job titles, generic expressions, national food references, abstract quality terms, and anonymized entities.
After the initial annotation phase and guideline updates, a second revision round was conducted to further improve consistency. 
In this phase, 48 sentences were revised, affecting only aspect phrase boundaries, while polarity and category assignments remained unchanged. 
Revisions were applied by one annotator and validated by the second, with remaining disagreements resolved through discussion.

%-----------------------------Crowd-----------------------------
\paragraph{Crowdworkers} For the dataset based on crowdworker annotations, we recruited 30 participants via Prolific\footnote{\scriptsize \href{https://www.prolific.com/}{https://www.prolific.com/}} to annotate 200 sentences each.
Participation was restricted to German-speaking individuals located in Germany, Austria, or Switzerland with at least secondary education.
%Participants who had taken part in related annotation studies were excluded.
Participants who had taken part in related annotation studies within this project were excluded to prevent overlap. 
Each participant was allowed to annotate only a single batch and could not participate in multiple studies (e.g., across the ACSA and TASD tasks).
Annotators were compensated at £9 per hour in line with Prolific’s recommendations. 
Each participant could submit only once, and unusually fast submissions were filtered using Prolific’s automated quality checks. 
The task was restricted to desktop devices to ensure consistent interaction with the annotation interface. 
Before annotation, participants completed a short questionnaire and provided informed consent via Google Forms.
Annotations were performed independently using Label Studio\footnote{\scriptsize \href{https://labelstud.io/}{https://labelstud.io/}} following detailed written guidelines and a short instructional video. 
% \subsection*{Pilot Study}
% A pilot study for the TASD task was conducted with five slots to validate the annotation setup. Although 15 participants registered, high dropout rates were observed: one participant timed out, nine withdrew early, and one discontinued after using an automatic translation tool. The pilot informed subsequent adjustments to task design and instructions.

%-----------------------------Students-----------------------------
\paragraph{Students} For this study, computer science-related students were recruited via the university network.
% \textbf{Students} were recruited via the university’s \textit{Versuchspersonenstunden} forum and personal contacts of the authors.
% All annotators had a background in Media Informatics and were compensated with 2 \textit{Versuchspersonenstunden}. 
Participants received written guidelines and an instructional video explaining the annotation procedure and use of Label Studio.
% The annotation was conducted between August~8 and September~3 2025.
Each batch of 200 texts was annotated independently by three students.
Final labels were derived using majority voting, requiring agreement on category, polarity, and aspect phrase. 
After removing the conflict label, sentences without remaining annotations were retained to ensure consistent dataset sizes across training sets.
%In the ACSA task, two annotators systematically misinterpreted the guidelines by assigning polarities to all categories. 
%These annotations were nevertheless retained and included in the final dataset.
Two annotators systematically misinterpreted the guidelines by assigning a sentiment label to every aspect category.
Instead of leaving non-mentioned categories unannotated, they labeled them as neutral (e.g., assigning neutral to Ambience and Price in a sentence that only expresses sentiment about Food and Service).
These annotations were retained in the final dataset, as they reflect a common and instructive source of error in annotation studies.
Before starting the annotation, participants completed a short questionnaire via Google Forms, which was identical for both students and crowdworkers to ensure consistent instructions. 
The questionnaire collected study information, obtained informed consent, and recorded participants’ prior annotation experience (ranging from no experience to professional level) and the annotation domains they had worked in (e.g., text, audio, video, image). 
The results, detailing experience level and type across students and crowdworkers, are provided in Appendix~\ref{App:questionaire_results}.
%-----------------------------LLM-----------------------------
\paragraph{LLMs} Following the methodology of \citet{Hellwig2025-sd}, we adapted their approach to annotate the training set using an LLM. 
For our experiments, we employed Gemma-3-27B with a temperature of 0.8, a context length of 4096 tokens, and five random seeds (0~–~4).
Based on the results reported by \citet{Hellwig2025-sd}, we injected 30 few-shot examples into the prompt. Given that configurations with 30 and 50 examples yielded similar performance in their study, we selected the smaller setting to reduce computational cost and annotation time.
To consolidate outputs across seeds, we applied a majority-voting procedure inspired by the self-consistency technique of \citet{wang_self-consistency_2023}, whereby an annotation was included in the final annotation if it appeared in the majority of seed predictions.

%-----------------------------Experts-----------------------------
\paragraph{Experts} Expert annotation was based on the original labels provided by \citet{hellwig_gerestaurant_2024}, which were transformed to match the revised labeling schema of our annotation interface. 
The expert annotator was a PhD student with prior ABSA experience and involvement in the original annotation as a curator.
For the TASD dataset, the expert reviewed all 1,000 texts using Label Studio’s review functionality, accepting, revising, or removing annotations per the updated guidelines.
In total, 1,333 annotations were accepted, 92 were revised, and 10 triplets were removed. 
Metadata tags were used to flag difficult or context-dependent cases.

% %-----------------------------Annotation Setup-----------------------------
% \subsection{Annotation Setup}
% %-----------------------------Interface-----------------------------
% The annotation was carried out using the web-based tool Label Studio, which was customized to support ABSA. 
% The annotation interface presents the sentence to be labeled at the top, followed by a structured labeling panel organized by five aspect categories.
% For each aspect category, annotators select one of four polarity options, implemented as color-coded buttons with keyboard shortcuts to support efficient annotation.
% An additional label "\texttt{Implizit}" (implicit) can be assigned in combination with an aspect label to indicate implicit aspect mentions. 
% The interface prevents invalid label combinations through warning messages.
% The interface further allows annotators to record metadata, such as annotation difficulty or insufficient context, and to provide optional free-text comments to justify their decisions.
% A link to the current annotation guidelines is included for quick reference.\footnote{Guidelines: Removed for double blind review}
% \begin{figure}[hbt!]
%     \centering
%     \includegraphics[width=0.483\textwidth]{images/Label_Interface_TASD.png}
%     \caption{Label interface for the TASD task in Label Studio with an annotated text example.}
%     \label{fig:annotation_setup_tasd}
% \end{figure}
%-----------------------------Guidelines-----------------------------

\paragraph{Annotation Guidelines} Two separate guideline sets were developed for TASD and ACSA, adapting the framework of \citet{hellwig_gerestaurant_2024}.\footnote{\scriptsize Guidelines:\,\hangindent=2.3em\hangafter=1 \href{https://github.com/NiklasDonhauser/absa-annotation-quality/tree/main/03_annotations/Guidelines}{https://github.com/NiklasDonhauser/\\absa-annotation-quality/tree/main/03\_annotations/Guidelines}} 
The TASD guidelines introduce ABSA, define aspect categories and polarity labels, and specify the annotation of aspect phrases, including the distinction between explicit and implicit mentions. 
They further outline constraints such as the requirement that sentiment must target the aspect and that only the first occurrence of an aspect phrase is annotated. 
Illustrative examples of complete aspect–category–sentiment triplets and practical instructions for using the annotation interface, including meta-tags and free-text comments, are also provided. 
The ACSA guidelines follow the same overall structure but omit the detailed specification of aspect phrases and adjust the interface instructions accordingly. 
Both guideline sets build on established ABSA annotation principles \citep{pontiki_semeval-2016_2016} and include concrete German-language examples.
In Appendix~\ref{App:label_interface}, we provide the layout of the annotation interface for both tasks (TASD and ACSA).

%% file: chapters/experiments.tex
The experiments evaluate the same ABSA subtasks considered in the annotation study, ACSA and TASD, to ensure direct comparability between annotation quality and downstream model performance.

\subsection{Baseline Methods}
To capture a broad range of methodological approaches, we consider classification-based architectures, text generation models, and LLMs. Due to the limited size of our datasets, constructing a separate development set was not feasible. 
Instead, we adopted the hyperparameter settings proposed by \citet{fehle_german_2025}, who applied the same models to the original GERestaurant dataset. 
For few-shot prompting experiments, we used Gemma 3 27B,\footnote{\scriptsize \href{https://huggingface.co/google/gemma-3-27b-it}{https://huggingface.co/google/gemma-3-27b-it}} following \citet{Hellwig2025-sd}, which also ensured consistency with the model used to generate the LLM dataset. 
%Since resource constraints prevented us from fine-tuning Gemma 3 27B, we employed LLaMA 8B\footnote{\url{https://huggingface.co/meta-llama/Llama-3.1-8B}} for instruction fine-tuned experiments, using the configuration and hyperparameters reported in \citet{fehle_german_2025} for the German restaurant domain.
Due to resource constraints, fine-tuning Gemma 3 27B was not feasible. 
Instead, we employed LLaMA 3.1 8B\footnote{\scriptsize\href{https://huggingface.co/meta-llama/Llama-3.1-8B}{https://huggingface.co/meta-llama/Llama-3.1-8B}} for instruction fine-tuned experiments. 
% Leider arxiv Paper zhou; zhou_comprehensive_2024_llama,
LLaMA-based models have demonstrated strong performance in fine-tuning settings for ABSA and related sentiment analysis tasks \citep{fehle_german_2025, fehle_leveraging_2026,  smid_llama-based_2024}. 
Building on these findings, we adopted the configuration and hyperparameters proposed by \citet{fehle_german_2025} for the German restaurant domain.
\paragraph{BERT-CLF}
Following \citet{fehle_aspect-based_2023}, we implement a multi-label classification model based on gbert-base.\footnote{\scriptsize\href{https://huggingface.co/deepset/gbert-base}{https://huggingface.co/deepset/gbert-base}} 
The model predicts aspect–sentiment pairs for the ACSA task, using a linear classification head on top of the [CLS] token representation from BERT. 
\paragraph{Hier-GCN}
The Hierarchical Graph Convolutional Network (Hier-GCN) \citep{Cai2020-km} combines contextual embeddings from gbert-base with graph convolutional layers to explicitly model dependencies between aspects and sentiments.
% Tokens are represented as nodes, and syntactic as well as semantic relations define the graph edges.
% The resulting graph representations are pooled and passed to a classifier to predict aspect–sentiment pairs in the ACSA task. 
\paragraph{Paraphrase}
The Paraphrase method \citep{zhang_method_paraphrase} treats the TASD task as a sequence-to-sequence text generation problem.
Using T5-base\footnote{\scriptsize\href{https://huggingface.co/google-t5/t5-base}{https://huggingface.co/google-t5/t5-base}} as the base model, the input sentence is reformulated into a natural-language template that explicitly encodes the target output structure.
% The model is trained to generate aspect–category–\linebreak polarity triplets directly from this reformulated input.
% This template-based approach ensures that outputs follow a predefined format and reduces structural errors common in free-form generation.  

\paragraph{MvP}
% The MvP approach \citep{gou_method_mvp} formulates the TASD task as a sequence-to-sequence text generation problem.
% Using T5-base as the underlying sequence-to-sequence model, the method generates aspect–category–polarity tuples directly from the input sentence by applying multiple prompt formulations (“views”). 
% The outputs from all views are then aggregated via a majority voting strategy to produce the final set of predictions, which helps reduce inconsistencies and improves robustness across diverse linguistic constructions.

The MvP approach \citep{gou_method_mvp} models TASD as a sequence-to-sequence generation task using T5-base. 
Multiple prompt formulations (“views”) are applied to generate aspect–category–polarity tuples, and predictions are aggregated via majority voting.

\paragraph{Few-Shot Prompting (Gemma FS)}
Few-shot prompting leverages LLMs via in-context learning to perform both ACSA and TASD tasks \citep{simmering_method_llm}.
We use the Gemma 3 27B model and provide 50 annotated examples directly in the prompt, randomly sampled from the corresponding training set.
Following \citet{Hellwig2025-sd}, who report the best performance with 50 examples, we adopt the same configuration to facilitate comparability with prior work.
The prompt template, adapted from \citet{gou_method_mvp}, is translated into German and tailored to the specific structure of each task.
Further details, including the instruction prompt and the structure of the examples, are provided in Appendix~\ref{App:prompts_for_fs_LLMS}.

\paragraph{Instruction-based Fine-Tuning (LLaMA FT)}
Instruction fine-tuning adapts a LLM to directly map input sentences to structured ABSA outputs \citep{smid_llama-based_2024}.
% We use LLaMA 3.1 8B and fine-tune it on task-specific datasets for both the ACSA and TASD tasks.
We follow the implementation of \citet{fehle_german_2025} and fine-tune LLaMA 3.1 8B on task-specific datasets for both the ACSA and TASD tasks. 
The same prompt template is used as in the few-shot setup to ensure consistency in task formulation.

%-----------------------------Evaluation Procedure-----------------------------
\subsection{Evaluation Procedure}
We evaluated annotation consistency, model performance, and statistical differences across datasets.
All experiments were conducted on a workstation with an NVIDIA Quadro RTX 6000 (24 \,GB GDDR6) GPU.

\paragraph{Inter-Annotator Agreement} For ACSA, we measured IAA using average pairwise micro-F1 and Krippendorff’s alpha \citep{krippendorff_computing_alpha} to account for chance agreement. 
For TASD, micro-F1 was used, following prior work \citep{chebolu_review_2023, pontiki_semeval-2016_2016}, as it captures overlap in aspect phrase spans.

\paragraph{Model Evaluation}
Models were assessed using micro-F1 scores, averaged over five runs with different random seeds, following previous work in German ABSA \citep{Hellwig2025-sd, fehle_german_2025}. 
To examine whether annotation sources influenced model performance, we conducted all statistical analyses separately for the ACSA and TASD tasks. 
For each task, we first assessed whether performance differed significantly between datasets when aggregating results across all models. 
In a second step, we analyzed each model individually to determine whether its performance was influenced by the dataset used, based on five independent runs per model.

Normality assumptions were evaluated using the Shapiro–Wilk test \citep{shapiro_analysis_1965}. Depending on the results, either parametric tests (repeated-measures ANOVA followed by paired t-tests) \citep{student_probable_1908, field_discovering_2012} or non-parametric alternatives (Friedman test followed by Wilcoxon signed-rank tests) \citep{friedman_use_1937, wilcoxon_individual_1992} were applied. Holm–Bonferroni correction was used to account for multiple comparisons \citep{holm_simple_1979}. Results were considered statistically significant at $p < 0.05$.

%% file: chapters/results_discussion.tex
This section reports results for the ACSA and TASD subtasks, including a comparative analysis of dataset variants, inter-annotator agreement, model performance across datasets, and cost and effort considerations.
%-----------------------------Comparativ Analysis-----------------------------
\subsection{Comparative Analysis of Dataset Variants}
% \paragraph{ACSA}
Across annotation approaches, category distributions for ACSA remain consistent with only minor shifts between datasets.
Variations are most pronounced for \texttt{GENERAL} and \texttt{FOOD}, while \texttt{PRICE} remains the most stable category across all annotations. 
Overall, these differences suggest that annotator type introduces small but systematic changes in category frequencies without substantially altering the overall distribution.

Since ACSA does not distinguish between explicit and implicit mentions, we additionally examine polarity distributions. 
Sentiment polarity is largely preserved across datasets, with only minor variation between annotation sources, indicating that polarity assignment is comparatively robust to annotator differences.

%-----------------------------Datasets Distribution-----------------------------
% \begin{figure}[hbt!]
%     \centering
%     \includegraphics[width=0.45\textwidth]{images/category_distribution_tasd.pdf}
%     \caption{Category distribution across datasets for the TASD datasets. The figure shows the number of annotated aspects per category across the different datasets. The total number of aspects per dataset is displayed at the end of each bar.}
%     \label{fig:tasd_category_distribution}
% \end{figure}

% \paragraph{TASD}
%Figure~\ref{fig:tasd_category_distribution} shows the category distributions for the TASD task. 
Compared to ACSA, the differences between annotation approaches are more pronounced for TASD.
The expert and LLM-annotated datasets consistently contain higher counts across categories, while student and crowdworker annotations yield noticeably fewer instances, particularly for less frequent categories such as \texttt{AMBIENCE} and \texttt{PRICE}.

Similar patterns emerge for explicit and implicit mentions as well as sentiment polarity. 
Expert and LLM datasets maintain higher and more balanced distributions, whereas student and crowdworker datasets exhibit systematic reductions across categories, polarity labels, and mention types. 
Overall, these results indicate that annotation expertise and automation strongly influence dataset size and class coverage for the more complex TASD task.
For a more detailed view of the datasets, including additional statistics, see Appendix~\ref{App:dataset_statistics}.
%-----------------------------IAA-----------------------------
\subsection{Interrater Agreement during Annotation}

The dataset creation process revealed approach-specific trade-offs: crowdsourcing and student annotations differed in reliability and timeliness, LLM-based annotation required substantial computational resources and may reflect training-data biases, and expert annotation achieved the highest quality but was limited in availability, leading to its restriction to TASD and a reduced ACSA formulation.

% The dataset creation process revealed approach-specific trade-offs. 
% Crowdsourcing and student annotations differed in reliability and timeliness, while the LLM-based dataset required substantial computational resources and may reflect training-data biases. 
% Expert annotations achieved the highest quality but were constrained by limited availability, resulting in their restriction to TASD with the phrase component removed for ACSA to reduce annotation effort.

% The creation of the datasets revealed distinct practical challenges across annotation approaches. 
% Crowdsourcing was relatively costly and produced annotations of fluctuating quality, while student annotations were more consistent but often delayed. 
% The LLM-based dataset required substantial computational resources and may reflect biases from the model’s training data. 
% Expert annotations yielded the highest quality but were limited by the availability of qualified annotators and were therefore restricted to TASD, with the phrase component removed for ACSA to reduce annotation effort.
Although we did not systematically evaluate the impact of guideline clarity or interface design, we assume that clear instructions and a carefully designed annotation interface likely contributed to reducing errors and improving annotation consistency overall.
However, the previously observed systematic misinterpretations indicate that certain aspects of the guidelines remained ambiguous, suggesting that not all sources of error can be mitigated through interface design alone, and that particular care is needed in formulating unambiguous annotation guidelines.
Questionnaire responses indicate that most students and crowdworkers had little or no prior annotation experience, with only a few reporting moderate or extensive experience. 
Prior work was mainly in text and image annotation, with smaller numbers having experience in audio, video, or multimodal tasks.
%Questionnaire data indicate that most student and crowdworker annotators had little or no prior experience. Among students, 13 reported no experience, 7 less than 10~hours, 4 between 10–50~hours, and 2 over 50~hours. Prior experience was primarily in text (6) and image (7) annotation, with 1 each for audio and multimodal tasks. Crowdworkers showed a similar profile: 10 had no experience, 7 less than 10~hours, 6 between 10–50~hours, 5 over 50~hours, 1 reported work experience, and 1 was unsure. Their annotation domains included text (9), image (11), audio (7), video (9), and multimodal (3), with multiple selections allowed.
Table~\ref{tab:iaa_total} presents a comparative overview of IAA for ACSA and TASD across all datasets, excluding the expert dataset due to the presence of only a single annotator.
\input{tables/iaa_full_2}
%-----------------------------IAA ACSA-----------------------------
\subsubsection{IAA on the ACSA Task}
The IAA results for the ACSA task show broadly consistent patterns across datasets. 
Crowdworker and student annotations achieve comparable agreement levels, reflecting the constrained annotation setup in which annotators select predefined category–polarity pairs. 
However, variability across annotation batches suggests that certain texts were more difficult or that annotators applied divergent interpretations. 
This effect is particularly evident in the student dataset, where unusually high variance in some batches indicates misinterpretation of the guidelines, inflating the overall agreement variance compared to the crowdworker dataset.
These observations underscore that annotation errors can occur even with clear instructions, highlighting the importance of carefully designed guidelines and interfaces \citep{klie_analyzing_2024}. 
In response, an additional warning was introduced in the crowdworker interface to mitigate similar issues.
In contrast, LLM-generated annotations show very high agreement.
Despite using a non-zero temperature to encourage output diversity, repeated prompting produced highly consistent annotations, explaining the strong IAA scores.
As noted by \citet{klie_analyzing_2024}, however, high agreement alone does not necessarily imply high annotation quality.
%-----------------------------IAA TASD-----------------------------
\subsubsection{IAA on the TASD Task}
For the TASD task, IAA differs more strongly across datasets, reflecting the increased annotation complexity. 
Unlike ACSA, annotators were required to freely select text spans, which substantially increased variability. 
Student annotations generally adhered to the guidelines, whereas crowdworker annotations often included overly long spans, partial sentence fragments, or inconsistent handling of implicit aspects.
Additional errors included splitting multi-word aspects into several single-token annotations, resulting in highly variable annotation quality. 
These issues occurred far less frequently in the student dataset and are reflected in the higher variance observed for crowdworker annotations.
As in ACSA, LLM-generated annotations exhibit very high agreement, despite the use of a non-zero temperature, due to the fixed prompting setup. 
Overall, IAA for TASD is notably lower than for ACSA, consistent with the added difficulty of aspect phrase extraction. 
This finding aligns with prior work \citep{monarch_human---loop_2021}, which shows that tasks challenging for human annotators also tend to be difficult for machine learning models.
%-----------------------------IAA GT-----------------------------
\subsubsection{IAA on the Ground Truth}
IAA for the ground truth TASD dataset is consistently high and clearly exceeds that of the student and crowdworker annotations. 
Agreement improves steadily across annotation batches, indicating that iterative discussions, calibration, and guideline refinements led to increasingly consistent annotations with low variance.
These findings highlight the importance of expert collaboration and regular feedback in producing reliable gold-standard annotations for complex tasks such as TASD, in line with prior work emphasizing the role of careful guideline design and calibration in improving annotation quality \citep{klie_analyzing_2024, fehle_german_2025}.
%-----------------------------Model Performacne-----------------------------
\subsection{Model Performance on the different Datasets}
This section analyzes model performance across the two ABSA subtasks: ACSA and TASD. 
We compare classical baselines and LLM-based approaches trained on datasets annotated by different annotator types.
%-----------------------------Performance ACSA-----------------------------
\input{tables/results}
\subsubsection{Performance on the ACSA Task}
Table~\ref{tab:results_f1_acsa} summarizes model performance on the ACSA task. 
Overall, models trained on expert-annotated data achieve the strongest results across nearly all approaches, although differences between datasets are relatively small. 
LLM-based models outperform classical baselines, with fine-tuned and few-shot LLMs on expert annotations achieving the strongest overall results, in line with prior findings by \citet{fehle_german_2025, fehle_leveraging_2026}.
An exception is the few-shot setting, where the student dataset yields the highest score, indicating that non-expert annotations can occasionally be competitive.

Across models, performance remains relatively stable regardless of the annotation source, suggesting that all datasets are broadly suitable for ACSA. 
Nevertheless, expert annotations consistently provide a small but reliable advantage when optimizing for performance.

Analysis by category and polarity shows that positive sentiment is easiest to predict, followed by negative, while neutral sentiment remains the most challenging. 
Performance drops are particularly pronounced for neutral polarity in the \texttt{AMBIENCE} category, whereas \texttt{FOOD} benefits from a higher number of neutral examples.

Pairwise tests reveal isolated significant differences for Hier-GCN (Experts vs.\ Students) and Gemma-FS (Experts vs.\ LLMs; LLMs vs.\ Students), without systematic effects over all datasets.

% stat tests:
% No significant differences were observed in the overall comparison across datasets. 
% In the per-model analysis, significant differences were found for Hier-GCN between Experts and Students (adjusted p = 0.0393), for Gemma-FS between Experts and LLMs (adjusted p = 0.0329), and between LLMs and Students (adjusted p = 0.0188). 
% LLaMA-FT and BERT-CLF showed no significant differences after Holm-Bonferroni correction.
%-----------------------------Performance TASD-----------------------------
\subsubsection{Performance on the TASD Task}
Table~\ref{tab:results_f1_tasd} summarizes TASD performance across datasets. 
As in ACSA, expert-annotated data yields the strongest results overall, with the highest scores for most models. 
The best performance is achieved by the fine-tuned LLM, which clearly outperforms all other approaches on the expert dataset. 
An exception is the few-shot LLM setting, where the LLM-annotated dataset performs slightly better, likely due to the shared underlying model.

Across models, the crowdworker dataset consistently results in the lowest performance, while student and LLM datasets show comparable results, occasionally outperforming each other depending on the model.
Compared to ACSA, performance differences between datasets are more pronounced for TASD, reflecting the increased task complexity and lower annotation agreement.

Category–polarity analysis follows similar patterns to ACSA: positive sentiment is easiest to predict, followed by negative, while neutral sentiment remains the most challenging. 
Performance drops are especially pronounced for infrequent classes such as \texttt{PRICE} and for neutral polarity, particularly in the \texttt{AMBIENCE} category.

Statistical testing reveals significant overall differences between Crowdworker and Student datasets, with further pairwise effects for Paraphrase and MvP.

% stat test:
% The overall comparison across datasets revealed that only Crowdworkers and Students reached statistical significance (adjusted p = 0.0106). 
% In the per-model analysis, most pairwise comparisons for Paraphrase and MvP remained significant after Holm–Bonferroni correction, with the only exception being the LLM vs. Student comparison for Paraphrase. 
% For Gemma-FS and LLaMA-FT, no pairwise differences reached significance.
%-----------------------------Cost-----------------------------
\subsection{Cost and Effort Analysis}

Creating the datasets involved varying levels of cost and effort. 
Crowdworker studies were completed within two days per study at a total cost of roughly $\pounds828 (\hat{=} \pounds0.41$ per three way   annotated example), including platform fees. 
Student annotations required several weeks while engagement relied on course credits. 
For LLM-based annotation, assuming $\$0.05-\$0.25$ per million tokens (MTok) for a self-hosted model~\cite{knoop2026-co} and a budget of $\approx$10{,}000 tokens per example, the upper-bound inference cost is at approximately $\$0.0025$ per example. 
By contrast, commercial frontier APIs such as GPT-5.2 Pro ($\$21/\$168$ per 1M input/output tokens) yield an estimated cost of $\approx\$0.36$ per example. 
Thus, self-hosted LLM annotation is substantially cheaper per example than crowd-based annotation, while frontier APIs approach similar cost levels.
Expert refinement required several hours; without pre-existing labels, large-scale expert annotation would be costly and difficult to scale due to limited availability.

%-----------------------------Summary-----------------------------
\subsection{Summary}
The discussion highlights key insights into dataset creation, annotation quality, and model performance for ABSA. 
Clear annotation guidelines and well-designed interfaces are crucial for reducing errors and improving consistency. 
While IAA is a useful indicator of task complexity and reliability, high agreement does not necessarily translate into superior model performance, as shown by LLM annotations, which achieved high IAA but only comparable performance to student-annotated data.
Each annotation strategy involves trade-offs: crowdsourcing can be costly and inconsistent, student annotations are slower, LLM-generated datasets require computational resources and may reflect training biases, and expert annotations are time-intensive but yield the highest quality. 
Across tasks, LLM-based models perform competitively, benefiting from large-scale pretraining, while expert annotations remain the most reliable basis for achieving peak performance.

Overall, these findings suggest that LLMs provide a fast and scalable annotation \citep{dietz_principles_2025, li_llms-as-judges_2024} alternative for ABSA, whereas expert annotation remains the gold standard when maximum accuracy and reliability are required.

%% file: tables/iaa_full_2.tex
\begin{table*}[ht]
\footnotesize
\centering
\begin{tabular}{lllll llll}
\toprule
\multicolumn{1}{c}{\textbf{}}  & \multicolumn{4}{c}{\textbf{ACSA}} & \multicolumn{4}{c}{\textbf{TASD}} \\
\cmidrule(lr){2-5}
\cmidrule(lr){6-9}
 & \textbf{GT} & \textbf{Crowd} & \textbf{Students} & \textbf{LLMs} & \textbf{GT} & \textbf{Crowd} & \textbf{Students} & \textbf{LLMs}  \\
\hline
\rowcolor{lightgrayrow} Batch 1 & 83.93 & 66.75$_{\pm 11.31}$ & 85.11$_{\pm 1.43}$ & 98.12$_{\pm 0.55}$ & 63.33 & 44.47$_{\pm 16.20}$ & 41.29$_{\pm 17.67}$ & 90.20$_{\pm 2.11}$ \\
Batch 2 & 88.38 & 84.57$_{\pm 1.43}$ & 81.55$_{\pm 1.01}$ & 96.46$_{\pm 1.10}$ & 70.25 & 61.55$_{\pm 6.41}$ & 63.81$_{\pm 2.34}$ & 87.66$_{\pm 2.82}$\\
\rowcolor{lightgrayrow} Batch 3 & 89.66 & 78.94$_{\pm 2.19}$ & 50.65$_{\pm 25.84}$ & 96.23$_{\pm 1.49}$ & 75.78 & 28.78$_{\pm 20.15}$ & 45.85$_{\pm 7.19}$ & 90.74$_{\pm 2.21}$ \\
Batch 4 & 88.25 & 84.24$_{\pm 1.60}$ & 52.03$_{\pm 27.54}$ & 97.32$_{\pm 1.27}$ & 74.41 & 19.91$_{\pm 21.97}$ & 45.71$_{\pm 12.52}$ & 89.30$_{\pm 1.95}$ \\
\rowcolor{lightgrayrow} Batch 5 & 85.78 & 83.54$_{\pm 2.64}$ & 81.56$_{\pm 1.51}$ & 97.86$_{\pm 0.69}$ & 76.95 & 26.33$_{\pm 26.49}$ & 55.26$_{\pm 9.64}$ & 92.95$_{\pm 1.19}$ \\
\midrule
\rowcolor{white}
\textbf{Overall} 
        & \textbf{87.22} & \textbf{78.95$_{\pm 2.27}$} & \textbf{63.38$_{\pm 16.75}$} & \textbf{97.20$_{\pm 0.88}$} 
        & \textbf{72.18} & \textbf{32.38$_{\pm 10.18}$} & \textbf{50.50$_{\pm 5.84}$} & \textbf{90.22$_{\pm 1.82}$} \\
\bottomrule
\end{tabular}
\caption{Batch-wise IAA (micro-F1) for ACSA and TASD across annotation groups. 
GT (ground truth) was annotated on the test split by two expert annotators. 
The remaining groups were annotated on the training split. 
For the experts group on the training split, only one annotator was available, so no IAA could be computed. 
Values denote per-batch mean $\pm$ standard deviation; standard deviation is not reported for GT.}
%\caption{Batch-wise IAA for ACSA and TASD across four datasets: Ground Truth (GT), Crowdworkers (Crowd), Students, and LLMs. Values show per-batch average Micro-F1 and fluctuations between multiple annotators as standard deviation (SD).}

\label{tab:iaa_total}
\end{table*}

%% file: tables/results.tex
% \begin{table}[ht]
% \centering
% \scriptsize
% \setlength{\tabcolsep}{1pt}
% \resizebox{0.49\textwidth}{!}{ 
% \begin{tabular}{
% l
% >{\centering\arraybackslash}m{1.0cm}
% >{\centering\arraybackslash}m{1.0cm}
% >{\centering\arraybackslash}m{1.0cm}
% >{\centering\arraybackslash}m{1.0cm}
% }
% \toprule
% \multicolumn{5}{c}{\textbf{Aspect Category Sentiment Analysis (ACSA)}} \\
% \toprule
% \textbf{Method} & 
% \makebox[0.9cm]{\textbf{Crowd}} &
% \makebox[0.9cm]{\textbf{Students}} &
% \makebox[0.9cm]{\textbf{LLMs}} &
% \makebox[0.9cm]{\textbf{Experts}} \\
% \midrule
% \rowcolor{lightgrayrow} BERT-CLF        & 76.99 & 77.81 & 77.44 & \textbf{78.26}	\\
% Hier-GCN        & 79.66 & 78.97 & 79.13 &  \textbf{79.78}\\
% \rowcolor{lightgrayrow} Gemma FS        & 86.03  & \textbf{86.43}  & 85.60  &  86.29 \\
% % 82.49 & 78.82 & 84.24 & 81.68 \\
% LLaMA FT        & 85.64 & 85.71 & 
% 84.85 &  \textbf{86.39} \\
% \midrule
% \multicolumn{5}{c}{\textbf{Target Aspect Sentiment Detection (TASD)}} \\
% \midrule
% \rowcolor{lightgrayrow} Paraphrase      & 52.77 & 57.33 & 57.37 & \textbf{61.65} \\
% MvP             & 51.29 & 56.83 & 60.65 & \textbf{64.01} \\
% \rowcolor{lightgrayrow} Gemma FS & 58.56 & 62.28 & \textbf{65.58}  & 63.38  \\
% LLaMA FT & 65.46 & 69.33 & 66.24 & \textbf{71.47} \\
% \bottomrule
% \end{tabular}
% }
% \caption{Micro-F1 scores for TASD and ACSA, averaged over five seeds across datasets. Bold indicates the highest values.}
% \label{tab:results_f1}
% \end{table}

\begin{table}[t]
\centering
\scriptsize
\setlength{\tabcolsep}{1pt}

\begin{subtable}[t]{\linewidth}
\centering
\resizebox{\textwidth}{!}{ 
\begin{tabular}{
l
>{\centering\arraybackslash}m{1.0cm}
>{\centering\arraybackslash}m{1.0cm}
>{\centering\arraybackslash}m{1.0cm}
>{\centering\arraybackslash}m{1.0cm}
}

\toprule
\textbf{Method} & 
\makebox[0.9cm]{\textbf{Crowd}} &
\makebox[0.9cm]{\textbf{Students}} &
\makebox[0.9cm]{\textbf{LLMs}} &
\makebox[0.9cm]{\textbf{Experts}} \\
\midrule
\rowcolor{lightgrayrow} BERT-CLF & 76.99 & 77.81 & 77.44 & \textbf{78.26} \\
Hier-GCN & 79.66 & 78.97 & 79.13 & \textbf{79.78} \\
\rowcolor{lightgrayrow} Gemma FS & 86.03 & \textbf{86.43} & 85.60 & 86.29 \\
LLaMA FT & 85.64 & 85.71 & 84.85 & \textbf{86.39} \\
\bottomrule
\end{tabular}
}
\caption{Aspect Category Sentiment Analysis (ACSA)}
\label{tab:results_f1_acsa}
\end{subtable}

\vspace{0.5em}

\begin{subtable}[t]{\linewidth}
\centering
\resizebox{\textwidth}{!}{ 
\begin{tabular}{
l
>{\centering\arraybackslash}m{1.0cm}
>{\centering\arraybackslash}m{1.0cm}
>{\centering\arraybackslash}m{1.0cm}
>{\centering\arraybackslash}m{1.0cm}
}
\toprule
\textbf{Method} & 
\makebox[0.9cm]{\textbf{Crowd}} &
\makebox[0.9cm]{\textbf{Students}} &
\makebox[0.9cm]{\textbf{LLMs}} &
\makebox[0.9cm]{\textbf{Experts}} \\
\midrule
\rowcolor{lightgrayrow} Paraphrase & 52.77 & 57.33 & 57.37 & \textbf{61.65} \\
MvP & 51.29 & 56.83 & 60.65 & \textbf{64.01} \\
\rowcolor{lightgrayrow} Gemma FS & 58.56 & 62.28 & \textbf{65.58} & 63.38 \\
LLaMA FT & 65.46 & 69.33 & 66.24 & \textbf{71.47} \\
\bottomrule
\end{tabular}
}
\caption{Target Aspect Sentiment Detection (TASD)}
\label{tab:results_f1_tasd}
\end{subtable}

\caption{Micro-F1 scores averaged over five seeds. Results are reported for ACSA and TASD across annotation sources (Crowd, Students, LLMs, Experts). Bold indicates the best performance per row.}
\label{tab:results_f1}
\end{table}

%% file: chapters/conclusion.tex
This study systematically analyzed the impact of annotation quality and annotator type on ABSA datasets and model performance.
We created a ground truth dataset annotated by two experts and four training datasets produced by crowdworkers, students, LLMs, and task experts, and evaluated SOTA models on ACSA and TASD, alongside IAA analyses.
Expert-annotated datasets consistently achieved the highest performance, with LLM-based models generally outperforming classical approaches. 
For ACSA, LLM annotations approached expert-level quality, while TASD performance was similar across crowdworker, student, and LLM datasets.
IAA was lowest for crowdworkers and students, highest for LLMs in ACSA, and more variable for TASD.
These findings highlight the importance of structured annotation guidelines and careful interface design. 
Expert annotations improve dataset quality but are time-intensive, whereas LLM-generated annotations offer a scalable alternative with competitive performance.

Future work includes evaluating models and annotations in other domains, combining LLM and expert annotations, increasing annotator numbers, exploring alternative aggregation strategies, and relaxing strict phrase boundaries to improve coverage and model robustness.

%% file: chapters/limitations.tex
% This study has several limitations.
% While crowd annotations enabled rapid data collection, they incur financial costs that scale with dataset size. 
% Student-based annotations were particularly time-intensive, as recruitment and task completion often spanned the entire one-week period. 
% LLM-generated annotations may reflect biases present in the model’s training data and require substantial computational resources, including high-memory GPUs, which can limit reproducibility. 
% Expert annotations yielded the highest-quality data but depend on scarce expertise and are difficult to scale. 
% Finally, all datasets and analyses are confined to the restaurant review domain, limiting the generalizability of our findings to other domains and languages.
This study has several limitations.
While crowd annotations enabled rapid data collection, they incur financial costs that scale with dataset size. 
Student-based annotations were particularly time-intensive, as recruitment and task completion often spanned the entire one-week period. 
It should be noted that students and crowdworkers are not inherently distinct groups. 
However, student annotators enable more controlled sampling with respect to demographic characteristics, while crowdworkers generally represent a more diverse but less controllable population.
LLM-generated annotations may reflect biases present in the model’s training data and require substantial computational resources, including high-memory GPUs, which can limit reproducibility. 
Furthermore, as the same LLM (Gemma 3 27B) was used for both annotation and few-shot inference, results on LLM-annotated data may be biased due to underlying model.
This should be considered when interpreting the performance of the few-shot prompting approach on the LLM dataset.
In addition, potential data contamination cannot be fully ruled out, as restaurant reviews are a widely used benchmark domain and may already be represented in the pretraining data of LLMs.
Expert annotations yielded the highest-quality data but depend on scarce expertise and are difficult to scale. 
Furthermore, despite clear instructions prohibiting external assistance, we cannot fully exclude the possibility that student or crowdworker annotators used LLMs as supportive tools or partially outsourced their work, which may have influenced annotation characteristics. 
Finally, all datasets and analyses are confined to the restaurant review domain, limiting the generalizability of our findings to other domains and languages.

%% file: chapters/ethical_considerations.tex
% This research was conducted without industrial funding or commercial sponsorship. 
We used OpenAI’s GPT-4.5 as a coding assistant to support implementation tasks and as a writing aid to improve clarity and formulation of the manuscript.
The dataset and its annotations are available upon request from the authors to ensure responsible academic use, while the Python code for data collection and preprocessing is publicly available on GitHub.\footnote{\scriptsize GitHub:\,\hangindent=2.3em\hangafter=1 \href{https://github.com/NiklasDonhauser/absa-annotation-quality}
{https://github.com/NiklasDonhauser/\\absa-annotation-quality}}

No demographic information was collected from crowdworkers or students, thereby minimizing privacy risks. 
All annotation procedures followed established data protection and ethical guidelines and were reviewed to prevent potential harms.
All participants signed an informed consent form permitting the use of their annotations for research purposes.
As with any annotated dataset, individual judgment and bias cannot be fully avoided. 
For the crowdworker, student, and LLM annotations, this was mitigated via majority voting. 
Because bias is particularly critical in the ground truth, two expert annotators independently annotated the data and resolved disagreements by consensus. 
Some residual subjectivity may remain.

%% file: appendix/prompts.tex
%-----------------------------Structure-----------------------------
\definecolor{bracketblue}{rgb}{0.25,0.5,0.75}
\lstdefinelanguage{PlainWithBlueBrackets}{}

\lstdefinestyle{bluebrackets}{
    language=PlainWithBlueBrackets,
    basicstyle=\ttfamily\scriptsize,
    frame=single,           
    numbers=none,           
    showstringspaces=false,
    breaklines=true,
    breakatwhitespace=false,
    breakindent=0pt,
    captionpos=b,
    tabsize=4,
    literate=
      % {[}{{{\color{bracketblue}[}}}1
      % {]}{{{\color{bracketblue}]}}}1
      % {aspect_category}{{{\color{highlightRed}aspect\_category}}}{13}
      % {'Positiv'}{{{\color{highlightGreen}'Positiv'}}}{7}
      % {'Negativ'}{{{\color{highlightGreen}'Negativ'}}}{7}
      % {'Neutral'}{{{\color{highlightGreen}'Neutral'}}}{7}
      % {'NULL'}{{{\color{highlightYellow}'NULL'}}}{6}
      % red category, green sentiment, yellow word
      {ä}{{\"a}}1 {Ä}{{\"A}}1
      {ö}{{\"o}}1 {Ö}{{\"O}}1
      {ü}{{\"u}}1 {Ü}{{\"U}}1
      {'Aspektkategorie'}{{{\color{highlightRed}'Aspektkategorie'}}}{15}
      {'Sentiment-Polarität'}{{{\color{highlightGreen}'Sentiment\-Polarit\"at'}}}{18}
      {'Aspektbegriff'}{{{\color{highlightYellow}'Aspektbegriff'}}}{13}
      {examples}{{{\color{highlightBlue}examples}}}{8}
      {ß}{{\ss}}1,
    aboveskip=10pt,
    belowskip=10pt,
    xleftmargin=10pt,
    xrightmargin=10pt,
    framexleftmargin=7pt,
    framexrightmargin=7pt,
    framesep=5pt
}

\lstset{style=bluebrackets}

\begin{lstlisting}[caption={Sample prompt for the TASD task showing few-shot examples before the task sentence.}]
Gemäß der folgenden Definition der Sentiment-Elemente:  

- Der 'Aspektbegriff' ist das genaue Wort oder die genaue Wortgruppe im Text, die eine spezifische Eigenschaft, ein Merkmal
oder einen Aspekt eines Produkts oder einer Dienstleistung darstellt, über die ein Nutzer eine Meinung äußern kann. Der Aspektbegriff kann 'NULL' sein, wenn der Aspekt implizit ist. 

- Die 'Aspektkategorie' bezieht sich auf die Kategorie, zu der der Aspekt gehört, und die verfügbaren Kategorien sind: [[aspect_category]]. 

- Die 'Sentiment-Polarität' beschreibt den Grad der Positivität, Negativität oder Neutralität, die in der Meinung zu einem bestimmten Aspekt oder Merkmal eines Produkts oder einer Dienstleistung ausgedrückt wird. Die verfügbaren Polaritäten 
sind: 'Positiv', 'Negativ' und 'Neutral'.  

Erkenne alle Sentiment-Elemente mit ihren jeweiligen Aspektbegriffen, Aspektkategorien und Sentiment-Polaritäten im folgenden Text im Format
[('Aspektkategorie', 'Sentiment-Polarität', 'Aspektbegriff'), ...].

[[examples]]
\end{lstlisting}
\label{app:prompt_general_tasd_llm}
% 
%
%-----------------------------Examples-----------------------------
%
%
\definecolor{bracketblue}{rgb}{0.25,0.5,0.75}
\lstdefinelanguage{PlainWithBlueBrackets}{}

\lstdefinestyle{bluebrackets}{
    language=PlainWithBlueBrackets,
    basicstyle=\ttfamily\scriptsize,
    frame=single,           
    numbers=none,           
    showstringspaces=false,
    breaklines=true,
    breakatwhitespace=false,
    breakindent=0pt,
    captionpos=b,
    tabsize=4,
    literate=
      {[}{{{\color{bracketblue}[}}}1
      {]}{{{\color{bracketblue}]}}}1
      {ä}{{\"a}}1 {Ä}{{\"A}}1
      {ö}{{\"o}}1 {Ö}{{\"O}}1
      {ü}{{\"u}}1 {Ü}{{\"U}}1
      {'Essen'}{{{\color{highlightRed}'Essen'}}}{7}
      {'Ambiente'}{{{\color{highlightRed}'Ambiente'}}}{10}
      {'Service'}{{{\color{highlightRed}'Service'}}}{10}
      {'Gesamteindruck'}{{{\color{highlightRed}'Gesamteindruck'}}}{16}
      {'Positiv'}{{{\color{highlightGreen}'Positiv'}}}{7}
      {'Negativ'}{{{\color{highlightGreen}'Negativ'}}}{7}
      {'Neutral'}{{{\color{highlightGreen}'Neutral'}}}{7}
      {'NULL'}{{{\color{highlightYellow}'NULL'}}}{7}
      {'Bier'}{{{\color{highlightYellow}'Bier'}}}{7}
      {'Plätze'}{{{\color{highlightYellow}'Pl\"atze'}}}{8}
      {'Köbes'}{{{\color{highlightYellow}'K\"obes'}}}{7}
      {ß}{{\ss}}1,
    aboveskip=10pt,
    belowskip=10pt,
    xleftmargin=10pt,
    xrightmargin=10pt,
    framexleftmargin=7pt,
    framexrightmargin=7pt,
    framesep=5pt
}

\lstset{style=bluebrackets}

\begin{lstlisting}[caption={Listing of 30 few-shot examples for the TASD prompt and the corresponding sentence to predict. For space reasons, only a subset is shown.}]
Text: Furztrocken.
Sentiment Elements: [('Essen', 'Negativ', 'NULL')]
Text: Die schönsten Plätze sind draußen an den Mauern der Kirche!
Sentiment Elements: [('Ambiente', 'Positiv', 'Plätze')]
Text: Das Bier schmeckt und die Köbes haben die liebenswerte witzige Art.
Sentiment Elements: [('Essen', 'Positiv', 'Bier'),
('Service', 'Positiv', 'Köbes')]
Text: Ich weiß nicht was das soll.
Sentiment Elements: [('Gesamteindruck', 'Negativ', 'NULL')]
Text: Vor dem Eingang war eine beeindruckende Schlange von wartenden Gästen.
Sentiment Elements: []
...
Text: Wir kommen gerne wieder!
Sentiment Elements: [('Gesamteindruck', 'Positiv', 'NULL')]
Text: [Sentence to predict]
Sentiment Elements:
\end{lstlisting}
\label{app:prompt_examples_tasd_llm}

%% file: appendix/annotation_example.tex
\newcolumntype{L}[1]{>{\raggedright\arraybackslash}p{#1}}
\newcolumntype{C}[1]{>{\centering\arraybackslash}p{#1}}

\begin{table}[ht]
\centering
\small
\setlength{\tabcolsep}{7pt}
\renewcommand{\arraystretch}{1.55}
\resizebox{1.0\columnwidth}{!}{%
\begin{tabular}{L{1.6cm} C{0.5cm} L{7cm} L{7.2cm}}
\toprule
\textbf{Category} & \textbf{ID} & \textbf{Extracted Triplets} \textit{(aspect, sentiment, target)} & \textbf{Sentence} \\
\midrule

\rowcolor{lightgrayrow} \textsc{Food} & 736 &
\texttt{[}\texttt{[}"essen", "positive", "Essen"\texttt{]},\newline\phantom{\texttt{[}}\texttt{[}"essen", "positive", "Wein"\texttt{]}\texttt{]} &
\textit{``Das Essen geschmackvoll, der Wein ein lecker Tröpfchen.''} \\

\textsc{Service} & 913 &
\texttt{[}\texttt{[}"service", "positive", "Personal"\texttt{]}\texttt{]} &
\textit{``Das Personal war freundlich und zuvorkommend.''} \\

\rowcolor{lightgrayrow}\textsc{General} & 832 &
\texttt{[}\texttt{[}"gesamteindruck", "negative", \textsc{null}\texttt{]}\texttt{]} &
\textit{``Wir würden diesen Ort nicht empfehlen.''} \\

\textsc{Ambience} & 11 &
\texttt{[}\texttt{[}"ambiente", "positive", "Brauhaus"\texttt{]}\texttt{]} &
\textit{``Ein tolles uriges Brauhaus mit viel Platz.''} \\

\rowcolor{lightgrayrow}\textsc{Price} & 303 &
\texttt{[}\texttt{[}"preis", "positive",\newline\phantom{\texttt{[}\texttt{[}}"Preis-/Leistungsverhältnis"\texttt{]}\texttt{]} &
\textit{``Fazit: Preis-/Leistungsverhältnis mehr als stimmig!''} \\

\bottomrule
\end{tabular}
}
\caption{Representative ground truth annotations covering all five aspect categories. Each entry lists the sample ID, the extracted opinion triplets in \textit{(aspect, sentiment, target)} format, and the corresponding source sentence.}
\label{tab:sentence_examples}
\end{table}

%% file: appendix/4_annotations_example.tex
% \begin{figure}[ht]
% \centering
% \includegraphics[width=\textwidth]{images/4_annotations.pdf}
% \caption{Example annotations of the same text by the four different annotations groups (crowdworkers, students, LLMs, and experts). For crowdworkers, students, and LLMs, multiple annotations per text were aggregated using majority voting, and the resulting labels are shown.}
% \label{fig:absa_example_4_annotations}
% \end{figure}

\begin{figure}[!ht]
\centering

\begin{subfigure}[t]{0.48\textwidth}
    \centering
    \includegraphics[width=\textwidth]{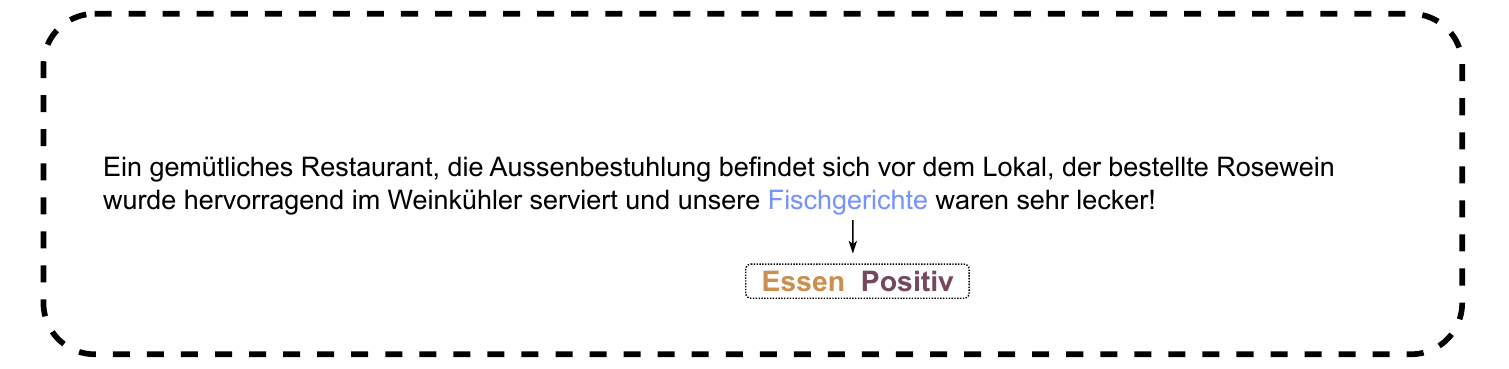}
    \caption{Crowdworkers}
\end{subfigure}
\hfill
\begin{subfigure}[t]{0.48\textwidth}
    \centering
    \includegraphics[width=\textwidth]{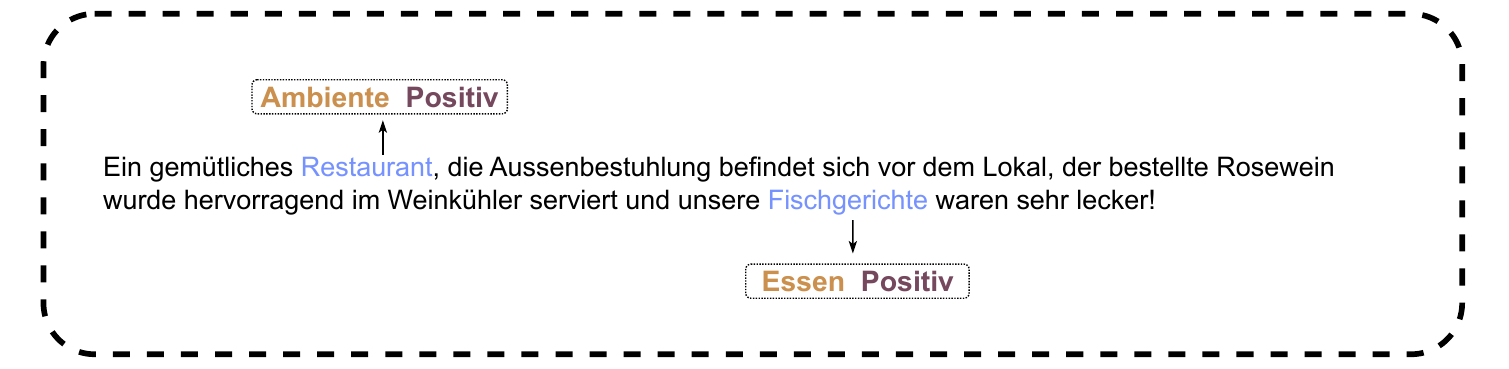}
    \caption{Students}
\end{subfigure}

\vspace{0.5em}

\begin{subfigure}[t]{0.48\textwidth}
    \centering
    \includegraphics[width=\textwidth]{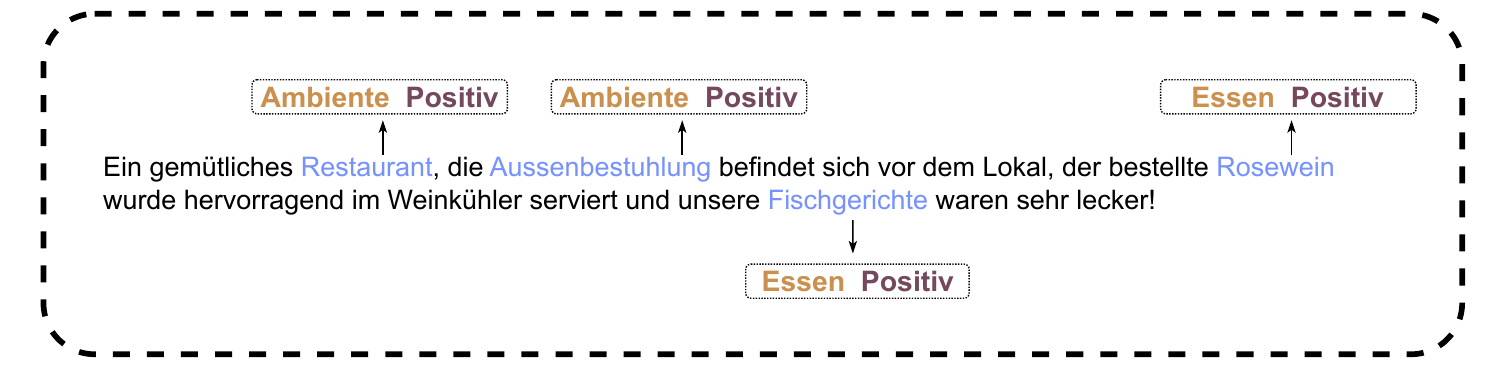}
    \caption{Large Language Models}
\end{subfigure}
\hfill
\begin{subfigure}[t]{0.48\textwidth}
    \centering
    \includegraphics[width=\textwidth]{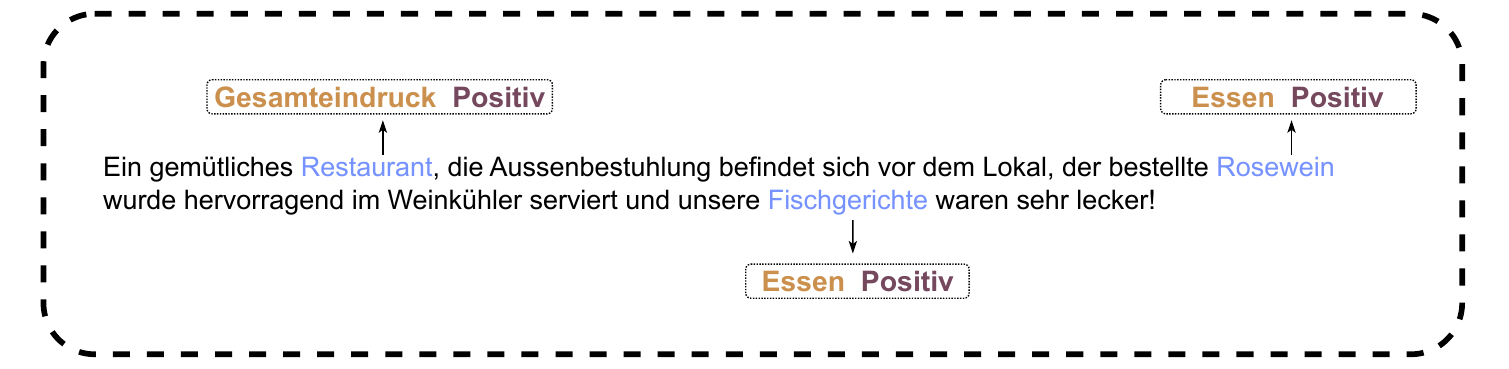}
    \caption{Experts}
\end{subfigure}

\caption{Example annotations of the same text by four groups (crowdworkers, students, LLMs, and experts). For crowdworkers, students, and LLMs, labels are aggregated via majority voting across multiple annotators.}
\label{fig:absa_example_4_annotations}
\end{figure}

%% file: appendix/sentiment_dataset_statistics.tex
\begin{table}[!ht]

\centering
\small
\setlength{\tabcolsep}{5pt}
\renewcommand{\arraystretch}{1.35}

%% ── Subtable 1: TASD ──────────────────────────────────────────────────────
\begin{subtable}{\linewidth}
\centering
\begin{tabular}{lrrrrrrrr}
\toprule
& \multicolumn{2}{c}{\textbf{Positive}}
& \multicolumn{2}{c}{\textbf{Negative}}
& \multicolumn{2}{c}{\textbf{Neutral}}
& \multicolumn{2}{c}{\textbf{Total}} \\
\cmidrule(lr){2-3}\cmidrule(lr){4-5}\cmidrule(lr){6-7}\cmidrule(lr){8-9}
\textbf{Annotator}
  & \textbf{Explicit} & \textbf{Implicit}
  & \textbf{Explicit} & \textbf{Implicit}
  & \textbf{Explicit} & \textbf{Implicit}
  & \textbf{Explicit} & \textbf{Implicit} \\
\midrule
\rowcolor{lightgrayrow}
Crowd    &   475 & 111 & 275 & 146 & 27 &  6 &   777 & 263 \\
Student  &   518 & 104 & 291 & 136 & 42 &  1 &   851 & 241 \\
\rowcolor{lightgrayrow}
LLM      &   596 & 177 & 336 & 221 & 64 & 11 &   996 & 409 \\
Experts  &   613 & 169 & 364 & 220 & 51 &  9 & 1,028 & 398 \\
\midrule
Test set &   526 & 146 & 340 & 199 & 48 & 16 &   914 & 361 \\
\bottomrule
\end{tabular}
\caption{Target Aspect Sentiment Detection (TASD)}
\end{subtable}

\vspace{0.9em}

%% ── Subtable 2: ACSA ──────────────────────────────────────────────────────
\begin{subtable}{\linewidth}
\centering
\begin{tabular}{lrrrr}
\toprule
\textbf{Annotator} & \textbf{Positive} & \textbf{Negative} & \textbf{Neutral} & \textbf{Total} \\
\midrule
\rowcolor{lightgrayrow}
Crowd    & 690 & 524 & 68 & 1,282 \\
Student  & 686 & 541 & 61 & 1,288 \\
\rowcolor{lightgrayrow}
LLM      & 670 & 515 & 75 & 1,260 \\
Experts  & 674 & 514 & 56 & 1,244 \\
\midrule
Test set & 671 & 515 & 56 & 1,242 \\
\bottomrule
\end{tabular}
\caption{Aspect Category Sentiment Analysis (ACSA)}
\end{subtable}

\caption{Distribution of sentiment labels (positive, negative, neutral) across datasets for the TASD and ACSA tasks, with a breakdown into explicit and implicit cases, reported for different annotation sources (crowd workers, students, LLMs, experts) and the test set. For the ACSA task, no explicit–implicit distinction is made; therefore, all values are reported under the explicit category.}
\label{tab:sentiment_distribution_overall_datasets}
\end{table}

%% file: appendix/category_dataset_statistics.tex
\begin{figure}[!ht]
    \centering
    % First subplot (top)
    \begin{subfigure}[b]{0.95\textwidth}
        \centering
        \includegraphics[width=\textwidth]{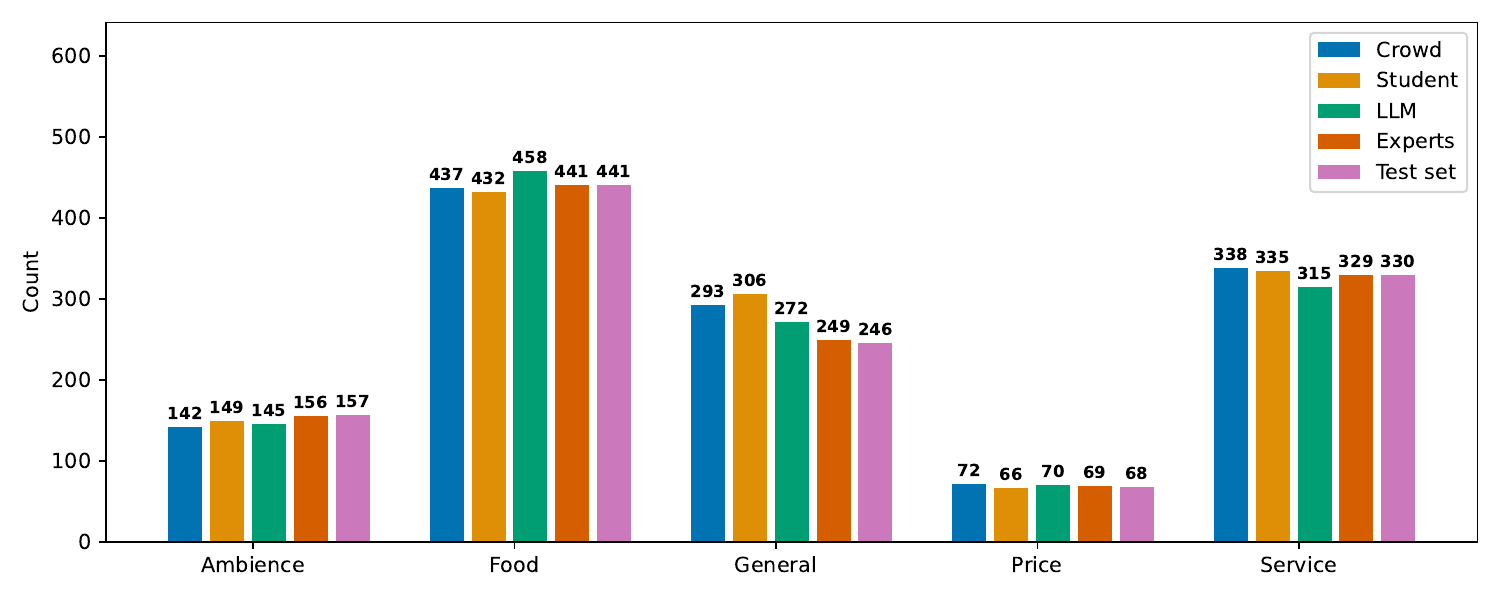}
        \caption{Aspect Category Sentiment Analysis (ACSA)}
        \label{fig:tasd_aspect_distribution}
    \end{subfigure}
    
    \vspace{0.5em} % small vertical space between figures
    
    % Second subplot (bottom)
    \begin{subfigure}[b]{0.95\textwidth}
        \centering
        \includegraphics[width=\textwidth]{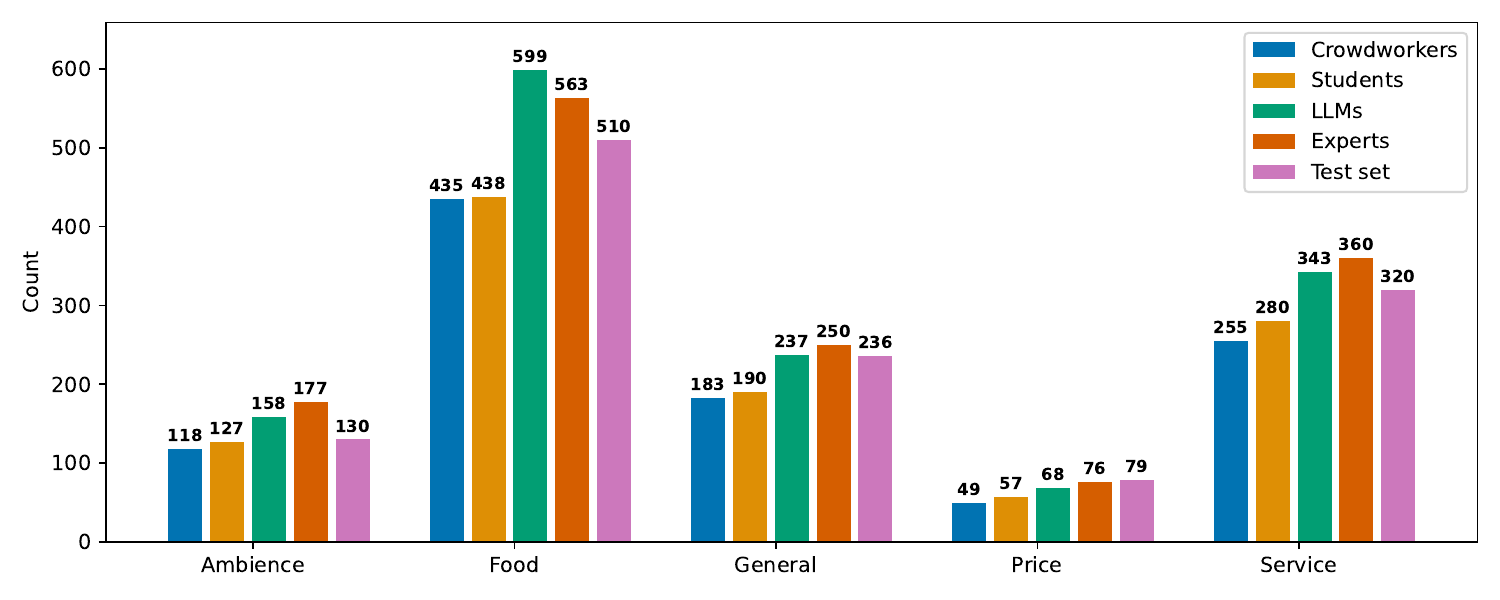}
        \caption{Target Aspect Sentiment Detection (TASD)}\label{fig:acsa_aspect_distribution}
        
    \end{subfigure}
    
    % Main caption for the entire figure
    \caption{Distribution of aspect categories across the five datasets for the TASD and ACSA tasks, reported for different annotation sources (crowdworkers, students, LLMs, experts) and the test set. Note that the test set contains fewer texts (924) compared to 1000 for the other datasets.}
    \label{fig:aspect_distribution}
\end{figure}

%% file: appendix/questionaire_results.tex
% \begin{table}[!ht]
% \centering
% \small
% \begin{subtable}[t]{0.4\textwidth}
% \centering
% \begin{tabular}{lcccc}
% \toprule
% & \multicolumn{2}{c}{\textbf{Students}} & \multicolumn{2}{c}{\textbf{Crowd}} \\
% \cmidrule(lr){2-3}\cmidrule(lr){4-5}
% \textbf{Experience Level} & \textbf{ACSA} & \textbf{TASD} & \textbf{ACSA} & \textbf{TASD} \\
% \midrule
% No experience & 9 & 8 & 6 & 4 \\
% less than 10 hours & 4 & 3 & 3 & 4 \\
% 10--50 hours & 2 & 2 & 2 & 4 \\
% more than 50 hours & 0 & 2 & 3 & 2 \\
% Work in field & 0 & 0 & 0 & 1 \\
% Don't know & 0 & 0 & 1 & 0 \\
% \bottomrule
% \end{tabular}
% \caption{Experience level distribution}
% \end{subtable}

% \hfill

% \begin{subtable}[t]{0.4\textwidth}
% \centering
% \begin{tabular}{lcccc}
% \toprule
% & \multicolumn{2}{c}{\textbf{Students}} & \multicolumn{2}{c}{\textbf{Crowd}} \\
% \cmidrule(lr){2-3}\cmidrule(lr){4-5}
% \textbf{Type of Experience} & \textbf{ACSA} & \textbf{TASD} & \textbf{ACSA} & \textbf{TASD} \\
% \midrule
% Text & 4 & 2 & 0 & 9 \\
% Image & 4 & 3 & 3 & 8 \\
% Audio & 1 & 0 & 3 & 4 \\
% Video & 0 & 0 & 4 & 5 \\
% Multimodal & 1 & 0 & 1 & 2 \\
% \bottomrule
% \end{tabular}
% \caption{Annotation modality distribution}
% \end{subtable}

% \caption{Comparison of annotators’ prior experience in terms of expertise levels and annotation modalities across student and crowdworker groups in the ACSA (n=15) and TASD (n=15) studies.}
% \label{tab:experience_type}
% \end{table}

\begin{table}[ht]
\centering
\small
\begin{subtable}[t]{0.45\linewidth}
\begin{tabular}{lcccc}
\toprule
& \multicolumn{2}{c}{Students} & \multicolumn{2}{c}{Crowd} \\
\cmidrule(lr){2-3}\cmidrule(lr){4-5}
Experience Level & ACSA & TASD & ACSA & TASD \\
\midrule
\rowcolor{lightgrayrow}No experience & 9 & 8 & 6 & 4 \\
less than 10 hours & 4 & 3 & 3 & 4\\
\rowcolor{lightgrayrow}10--50 hours & 2 & 2 & 2 & 4\\
more than 50 hours & 0 & 2 & 3 & 2\\
\rowcolor{lightgrayrow}Work in field & 0 & 0 & 0 & 1\\
Don't know & 0 & 0 & 1 & 0\\
\bottomrule
\end{tabular}
\caption{Experience level distribution}
\end{subtable}
\hfill
\begin{subtable}[t]{0.45\linewidth}
\begin{tabular}{lcccc}
\toprule
& \multicolumn{2}{c}{Students} & \multicolumn{2}{c}{Crowd} \\
\cmidrule(lr){2-3}\cmidrule(lr){4-5}
Type of Experience & ACSA & TASD & ACSA & TASD \\
\midrule
\rowcolor{lightgrayrow}Text & 4 & 2 & 0 & 9\\
Image & 4 & 3 & 3 & 8\\
\rowcolor{lightgrayrow}Audio & 1 & 0 & 3 & 4\\
Video & 0 & 0 & 4 & 5\\
\rowcolor{lightgrayrow}Multimodal & 1 & 0 & 1 & 2\\
Miscellaneous & 0 & 0 & 0 & 0 \\
\bottomrule
\end{tabular}
\caption{Annotation modality distribution}
\end{subtable}

\caption{Comparison of annotators’ prior experience in terms of expertise levels and annotation modalities across student and crowdworker groups in the ACSA (n=15) and TASD (n=15) studies.}
\label{tab:experience_type}
\end{table}

%% file: appendix/label_interface.tex
\begin{figure}[ht]
    \centering
    
    % ACSA subplot (top)
    \begin{subfigure}[b]{0.9\linewidth}
        \centering
        \fbox{\includegraphics[width=\textwidth]{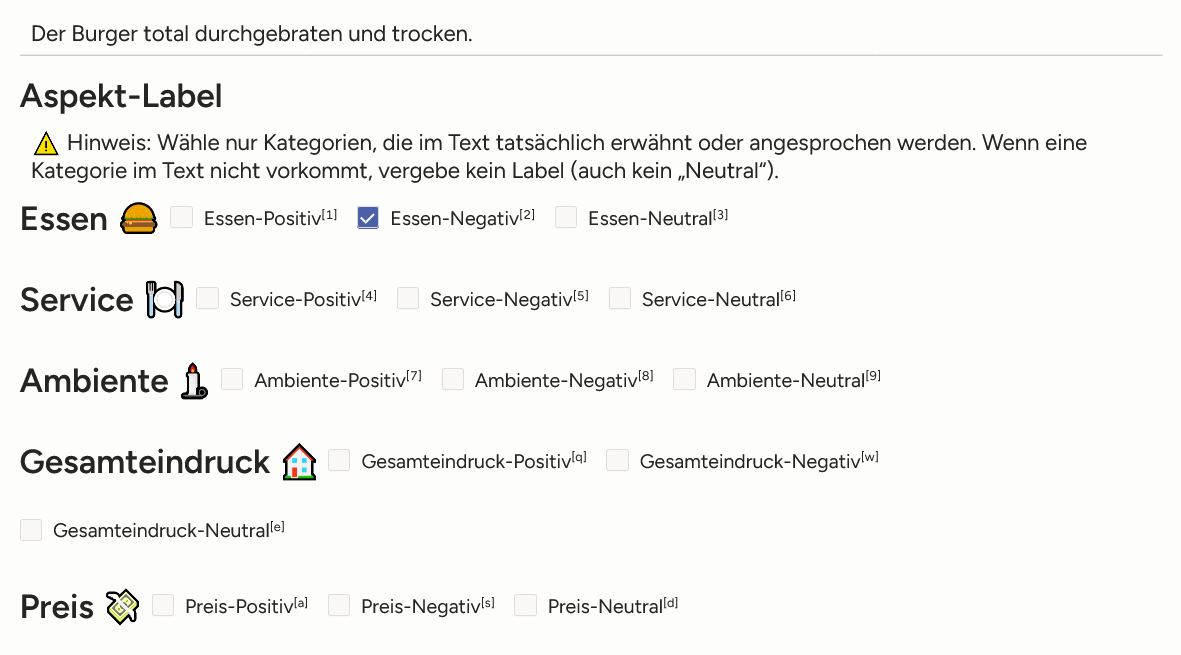}}
        \caption{Label interface for the ACSA task in Label Studio.}
        \label{fig:label_interface_acsa}
    \end{subfigure}
    
    \vspace{0.5em} % optional vertical space between images
    
    % TASD subplot (bottom)
    \begin{subfigure}[b]{0.9\linewidth}
        \centering
        \fbox{\includegraphics[width=\textwidth]{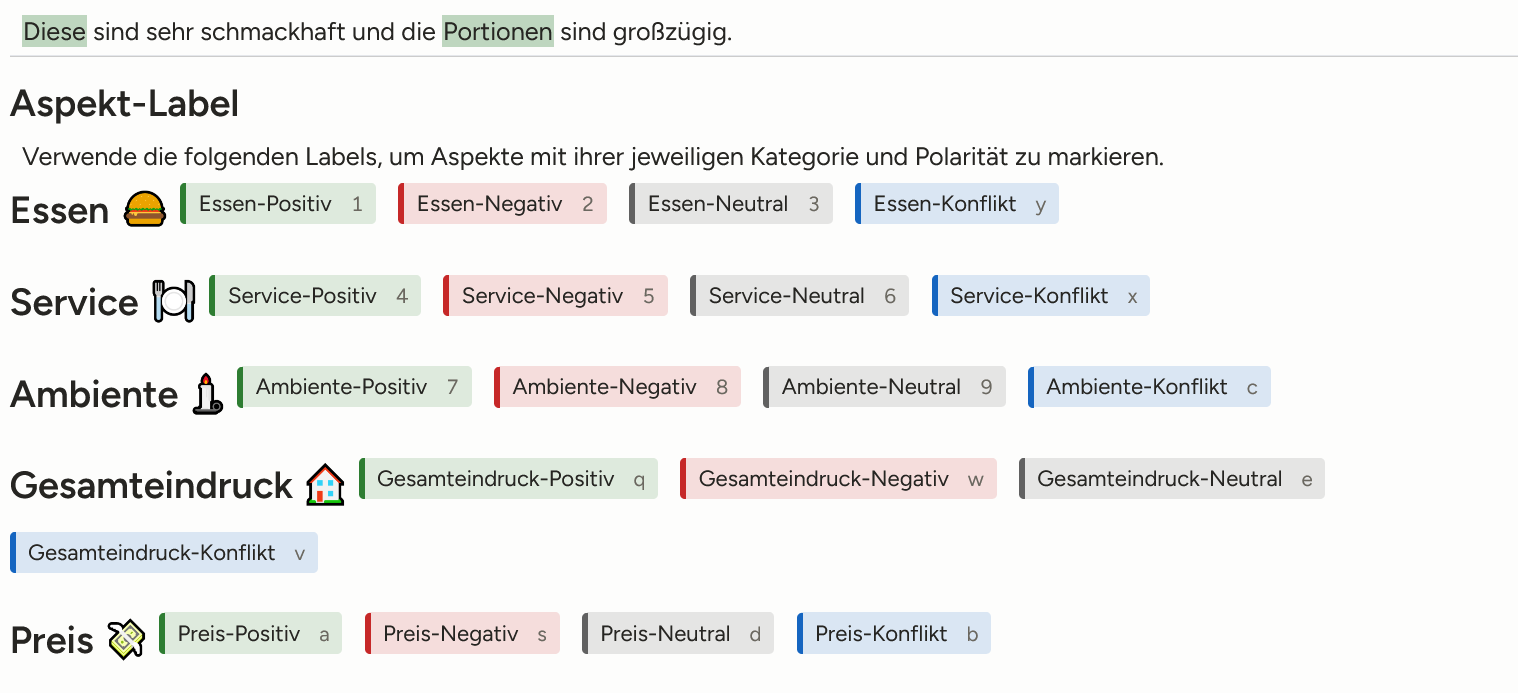}}
        \caption{Label interface for the TASD task in Label Studio.}
        \label{fig:label_interface_tasd}
    \end{subfigure}
    
    \caption{Label interfaces used for the ACSA and TASD annotation tasks. While not shown in the screenshots, both interfaces also included two meta tags (one for missing context and one to indicate difficult annotations) and a free-text field for annotator comments.}
    \label{fig:label_interfaces}
\end{figure}